%% file: MAIN.tex
\documentclass[letterpaper, preprint, paper,11pt]{AAS}	

\usepackage{bm}
\usepackage{amsmath}
\usepackage[colorlinks=true, pdfstartview=FitV, linkcolor=black, citecolor= black, urlcolor= black]{hyperref}
\usepackage{overcite}
\usepackage{footnpag}			      	
\usepackage{comment}
\usepackage{array}
\input{00_settings}

\newcommand{\mbf}{\mathbf}

\PaperNumber{25-793}

\begin{document}


\title{A Data-Driven Approach to Estimate LEO Orbit Capacity Models}

\author{Braden Stock\thanks{Undergraduate Research Assistant, Department of Aerospace Engineering, Iowa State University, Ames, IA 50011, USA. Email: bstock15@iastate.edu},
Maddox McCarthy\thanks{Undergraduate Research Assistant, Department of Aerospace Engineering, Iowa State University, Ames, IA 50011, USA. Email: madd0x@iastate.edu},  
Simone Servadio\thanks{Assistant Professor, Department of Aerospace Engineering, Iowa State University, Ames, IA 50011, USA. Email: servadio@iastate.edu},
}

\maketitle{}

\begin{abstract}
Utilizing the Sparse Identification of Nonlinear Dynamics algorithm (SINDy) and Long Short-Term Memory Recurrent Neural Networks (LSTM), the population of resident space objects, divided into Active, Derelict, and Debris, in LEO can be accurately modeled to predict future satellite and debris propagation. This proposed approach makes use of a data set coming from a computational expensive high-fidelity model, the MOCAT-MC, to provide a light, low-fidelity counterpart that provides accurate forecasting in a shorter time frame. 
\end{abstract}

\input{01_introduction}

\input{02_problem}

\input{03_method}

\input{04_results}
\input{05_significance}

\section*{Acknowledgments}
The authors wish to acknowledge the support of this work through the Iowa NASA EPSCoR Rapid Research Response (R3), grant number 80NSSC24M0124. The authors also wish to acknowledge the support of the MIT MOCAT development team.

\bibliographystyle{AAS_publication}   
\bibliography{references}   

\end{document}

%% file: 00_settings.tex
\usepackage{float}

\usepackage{nicefrac}

\usepackage{multirow}

\usepackage{cleveref}

\crefname{equation}{}{}

\usepackage{arydshln,leftidx,mathtools}

\usepackage{booktabs}
\usepackage{siunitx}

\usepackage{stackengine}
\stackMath

\usepackage{breqn}

\usepackage{bm}

\usepackage{diagbox}

\usepackage{tablefootnote}
\usepackage{threeparttable}

\usepackage{xcolor}

\newcommand{\RNum}[1]{\uppercase\expandafter{\romannumeral #1\relax}}

\usepackage{enumitem}
\usepackage{makecell}
\usepackage{textcomp}

\usepackage{tablefootnote}
\usepackage{threeparttable}

\usepackage{subcaption}

%% file: 01_introduction.tex
\section{Introduction}
\label{sec:introduction}

As governments and corporations continue to operate in Low Earth Orbit (LEO), the collision risk of objects increases. The Idirium-Cosmos collision in 2009 [\citen{wang2010analysis}] and the Chinese anti-satellite test on Fengyun-1c in 2007 [\citen{johnson2008characteristics}] resulted in 4000 pieces of trackable debris being created, posing a threat to operations in the near-earth environment. The Massachusetts Institute of Technology (MIT) Aerodynamics, Space, Robotics, and Control Laboratory has developed the Massachusetts Institute of Technology Orbital Capacity Assessment Tool (MOCAT) for further research into this topic. The MOCAT calculates active satellite (S), derelict satellite (D), and debris object (N) population in LEO over time. Within the MOCAT, the LEO environment is separated into 36 different atmospheric shells, starting at an altitude of 200 km and ending at an altitude of 2000 km, each with an equal altitude thickness of 50 km. The MOCAT currently has two different models for calculating satellite population changes over time: the Monte Carlo Model (MOCAT-MC) [\citen{jang2025new}] and the Source-Sink Evolutionary Model (MOCAT-SSEM) [\citen{d2023novel,gusmini2024effects}]. The Monte Carlo model has high fidelity but requires a longer computation time, while the Source-Sink model has lower fidelity but also has a lower computation time due to simplifications. These simplifications, however, hurt the accuracy of the model [\citen{d2023novel}]. The present work looks into the application of machine learning and data analysis, e.g., the SINDy and LSTM methods, as a means of improving and potentially replacing the performance of the MOCAT-SSEM in the future, to get a more accurate forecasting than the SSEM counterpart.

Sparse Identification of Nonlinear Dynamics (SINDy) is an open-source algorithm that utilizes sparse regression to derive differential equations to represent a system [\citen{brunton2016sparse}]. The algorithm works by receiving noisy data, building a nonlinear time series library, and then applying sparse regression to the library. On the other hand, the Long-Short-Term Memory (LSTM) networks [\citen{van2020review}] are a form of Recurrent Neural Networks (RNN) [\citen{grossberg2013recurrent}] that are better able to predict long-term trends in time series data in addition to shorter-term trends. LSTM networks address the vanishing gradient issue that occurs when using a normal RNN network by incorporating previous step data from both the immediate previous steps and steps further back in the time series relationship [\citen{klinkachornevaluating}]. By doing this, after training on a large enough sample data set, the LSTM network can accurately predict data at future time steps. By correctly implementing this type of network into the MOCAT, the MOCAT-SSEM model can be replaced, and the predictive capabilities of the model can be made more accurate and flexible compared to the Source-Sink model.

In this study, we apply the SINDy algorithm and an LSTM recurrent neural network model to simulate and predict data from the MOCAT-MC to verify the different models' applicability in predicting LEO satellite dynamics. This will aid the MOCAT project in its mission of risk assessment with the goal of providing viable replacement solutions for the MOCAT-SSEM and enhancing the LEO orbit capacity [\citen{servadio2024risk,servadio2024threat}] assessment.

%% file: 02_problem.tex
\section{Problem Statement}
\label{sec:problem}
Currently, the Source-Sink model of MOCAT, based on a set of coupled ordinary differential equations, has been derived by looking at empirically calculated parameters and coefficients [\citen{ashley2024parameters}], which are susceptible to initial conditions and by the specific application of interest. This model considers the subdivision of the LEO region into orbital altitude shells, assumed
to be spherical, and takes into account three species of ASOs, namely, active satellites (S), capable of performing collision avoidance
maneuvers; intact derelict satellites (D), which include disabled satellites and inactive satellites that fail to meet the PMD guidelines;
and debris. Because the proposed model is multishell, the LEO is assumed to be divided into many altitude shells; therefore, the resident space objects (RSOs) population is computed for each altitude shell. The differential equations used in the MOCAT-SSEM model to calculate the population change in a shell accounting only for the population belonging in that specific shell, with the exception of debris and derelict satellites falling from the current shell and falling from an above shell [\citen{d2023novel}]. Shown below are the differential equations used within the MOCAT-SSEM that currently propagate the LEO population with low fidelity. 
\begin{equation}
\dot{S}^{(i)} = \lambda^{(i)} - \frac{S^{(i)}}{TOF} - \alpha_a \phi_{SS} {S^{(i)}}^{2} - (\delta + \alpha) \phi_{SD} S^{(i)} D^{(i)} - (\delta + \alpha) \phi_{SN} S^{(i)} N^{(i)} \label{Eq. 1}
\end{equation}
\begin{equation}
\dot{D}^{(i)} = \frac{(1-PMD) S^{(i)}}{TOF} - \delta S^{(i)}D^{(i)} + \delta S^{(i)} N^{(i)} - \phi_{DD} {D^{(i)}}^{2} - \phi_{DN} D^{(i)} N^{(i)} + {\dot{F}_{d,D}}^{(i + 1)} - {\dot{F}_{d,D}^{(i)}} \label{Eq. 2}
\end{equation}
\begin{equation}
\begin{split}
\dot{N}^{(i)} &= n_{f,SS} \alpha_a \phi_{SS} {S^{(i)}}^{2} + n_{f,SD} \alpha \phi_{SD} S^{(i)} D^{(i)} + n_{f,SN} \alpha \phi_{SN} S^{(i)} N^{(i)} \\
&+ n_{f,DD} \phi_{DD} {D^{(i)}}^{2} + n_{f,DN} \phi_{DN} D^{(i)} N^{(i)} + n_{f,NN} \phi_{NN} {N^{(i)}}^{2} \\
&+ {\dot{F}_{d,N}}^{(i + 1)} - {\dot{F}_{d,N}}^{(i)} \label{Eq. 3}
\end{split}
\end{equation}
Where superscript $i$ corresponds to orbital shell $i=1,\dots,36$. Terms $S^{(i)}$, $D^{(i)}$, and $N^{(i)}$ correspond to the population of active satellites, derelict satellites, and debris, respectively, in shell $i$. Launch rate of new satellites into a given orbital shell is denoted by $\lambda^{(i)}$. Number of new fragments generated by collisions between objects are denoted by $\eta_{f,SS}$, $\eta_{f,SD}$, $\eta_{f,SN}$, $\eta_{f,DD}$, $\eta_{f,DN}$, and $\eta_{f,NN}$, where the subscript indicates which object types are involved in collision, and the values for each term is given by the NASA standard collision breakup model (EVOLVE 4.0) [\citen{NASA_breakup}]. Time of flight for active satellite operational lifespan is averaged at 5 years and represented as $TOF$. The parameterized probability of successful post-mission disposal is represented by term $PMD$ and valued at 0.95, based on empirical observations [\citen{JASON}]. Object orbit decay from the shell above ($i+1$) into the current shell ($i$) due to atmospheric drag are represented by $\Dot{F}_{d,D}^{(i+1)}$ and $\Dot{F}_{d,N}^{(i+1)}$, while object orbit decay from the current shell ($i$) into the shell below ($i-1$) due to atmospheric drag are represented by $\Dot{F}_{d,D}^{(i)}$ and $\Dot{F}_{d,N}^{(i)}$. Atmospheric drag is determined by a standard drag equation with ballistic coefficients, cross-sectional area, and velocity for $D$ and $N$ averaged from the initial population data set. Atmospheric density is either determined from a precalculated Jacchia-Bowman 2008 atmospheric density table or from a simple static exponential atmospheric density calculation. It is assumed that active control elements on board functional satellites prevent active satellites from experiencing orbital decay. 

This set of ODEs is repeated for each shell that divides the LEO space environment, for a total of 36 shells with a 50 km width each, starting at 200 km of altitude and ending at 2000 km. The final results are a system of 108 equations, with strong correlations among shells with their lower and upper bounds. 
\begin{figure}[H]
\centering
  \includegraphics[scale=0.50]{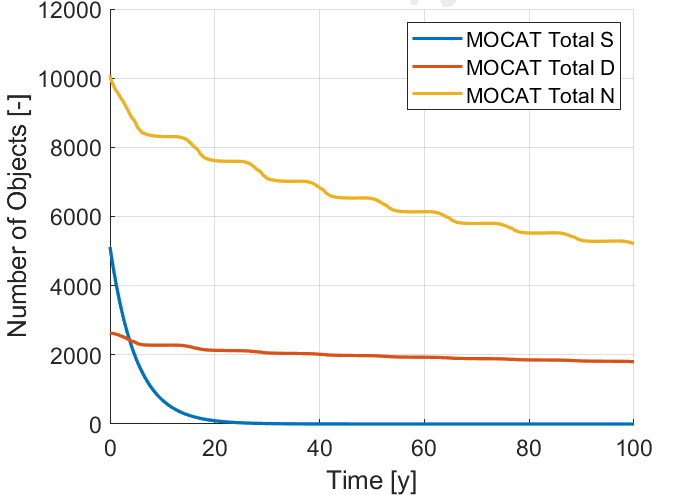}
  \caption{Diagram of MOCAT-SSEM for the null-launch case}
  \label{fig:ssem} 
\end{figure}
Figure \ref{fig:ssem} shows the total sum of active satellites, derelicts, and debris calculated via the SSEM model in the case when no active new launches happen in the future (null launch rate case). As expected, the S line declines to 0 in a few years, while derelict and debris continue orbiting higher altitude shells. Moreover, a large oscillatory behavior has been noted in lower shells. In shells 1-10, the Jacchia–Bowman 2008 (JB2008) thermospheric density model [\citen{bowman2008new}]  is a major factor in the dynamical system. This approximation changes the atmospheric density with the solar cycle, i.e., approximately every 11 years, resulting in the oscillations shown by the data.

\subsection{The Dataset}
\begin{figure}[h]
    \centering
    \includegraphics[width=1.00\textwidth]{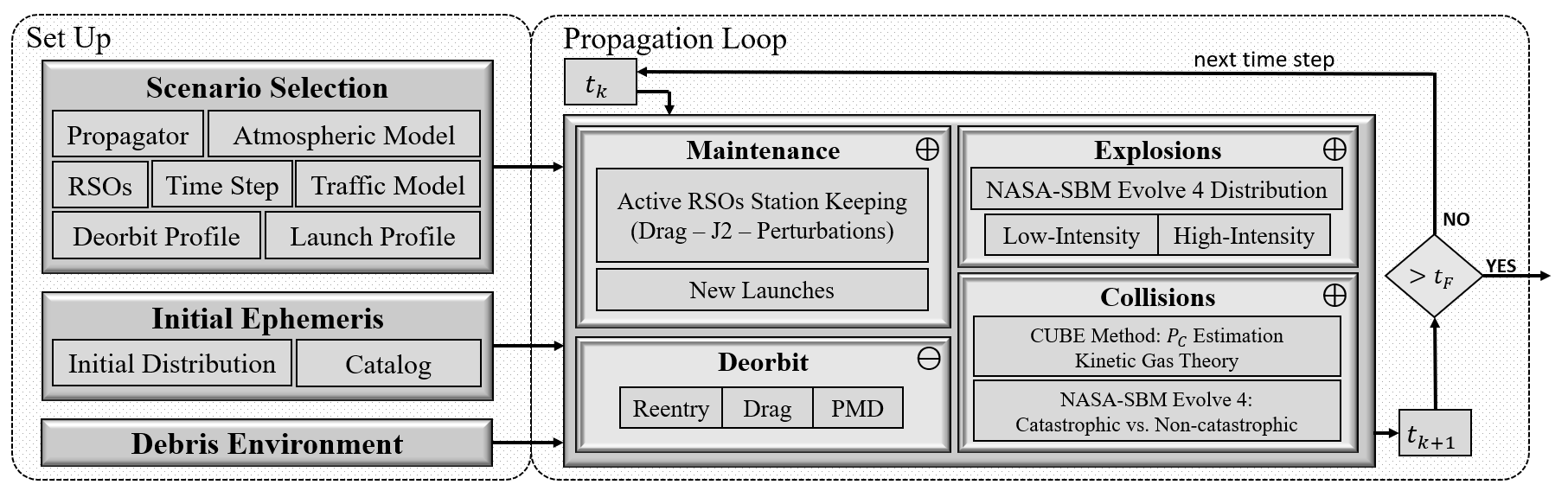}
    \caption{MOCAT-MC Architecture}
    \label{fig:mocat}
\end{figure}

\begin{figure}[!ht]
    \centering
    \includegraphics[width=0.49\linewidth]{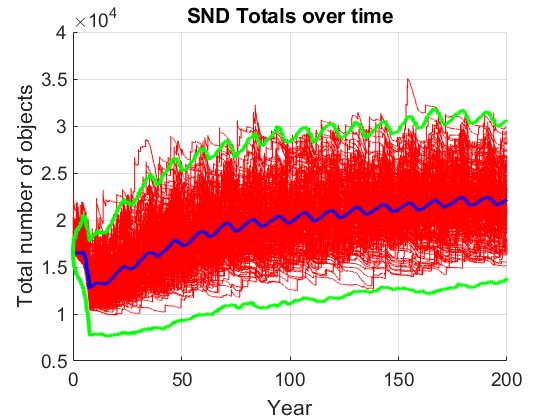}
    \includegraphics[width=0.49\linewidth]{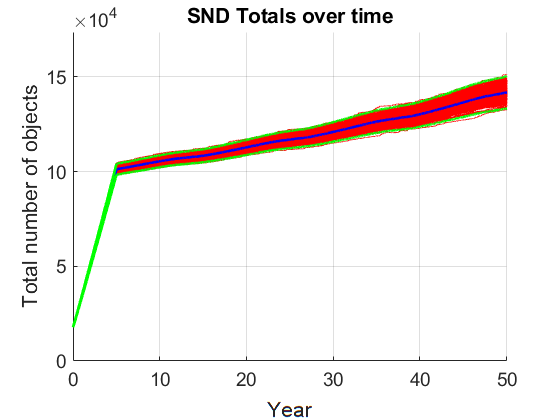}
    \caption{Evolution of S, N, D, For All Shells With Total Mean (Blue), 3$\sigma$ Deviation (Green), and Monte-Carlo Simulations (Red) for Null Launches (Left) and Constant Launches (Right).}
    \label{fig:SND_Tot}
\end{figure}

The dataset used to train the LSTM network and the SINDy least square optimization algorithm comes from the high fidelity propagation version of the MIT Orbit Capacity Toll, which propagates each single RSO singularly in a Monte Carlo sense: MOCAT-MC. The MOCAT-MC propagates, with high fidelity, the LEO space population modeling station keeping, collisions, explosions, and interactions according to the most current model, such as the NASA Standard Break-Up Model EVOLVE 4.0. The architecture of MOCAT-MC is reported in Figure \ref{fig:mocat}, and it has been used in recent years to support the need for orbital policy, conjunction assessment, launch rate identification, collision risk evaluation, and active debris removal [\citen{simha2025optimal}]. 
In the current version of the orbit capacity tool, whenever a collision is simulated, debris is generated according to the NASA model and gets associated with a Two Line Element (TLE). The LSTM and SiNDy will try to emulate the results of the high fidelity model by providing a new low fidelity source-sink evolutionary model to replace MOCAT-SSEM. Therefore, the MOCAT-MC is run multiple times to obtain a set of many (equally probable) possible revelations of the future population in the LEO space. The time evolution of the population is stored as a series of TLE for each time step, which have been grouped in their corresponding orbital shell. The results of the data extraction process for all thirty-six orbital shells are shown in Fig. \ref{fig:SND_Tot}. Each red line represents a different MOCAT-MC simulation,  with mean in blue and 3$\sigma$ standard deviation in red.

The SINDy and LSTM methods have multiple parameters that require tuning to obtain accurate results. With SINDy, the sampling rate ratio and sparsification tolerance are the major parameters that control the produced dynamics. With the LSTM model, the number of hidden layers, nodes per hidden layer, training data ratio, activation function selection, loss function selection, look-back distance, and number of training epochs are the major parameters that control the accuracy of the predicted data. However, the entire available data set is too large to optimize these parameters efficiently, and it requires prior data management and analysis. 

To obtain the orbital capacity training and baseline comparison data, we used the dataset from the MOCAT-MC model, 4,000 simulations of satellite populations, each simulating a different possible future LEO population due to the stochastic nature of the Monte-Carlo simulation. Each simulation ran for a time period of 100 years with a time step of 5 days.  The data from each simulation was then packaged into a MATLAB data file for analysis. For the preliminary results outlined in this abstract, the MOCAT data that was used to train both the SINDy and LSTM models was the average of all 4,000 simulations, extracted in terms of mean and standard deviation. This average was taken over each timestep and over each type of object for each atmospheric shell. More detailed results will be obtained for each simulation individually.

%% file: 03_method.tex
\section{The Sparse Identification of Nonlinear Dynamics (SINDy) Approach} \label{sec:sindy}

The data-file averaged results provide the average of active satellites, derelicts, and debris per shell over time. These mean values can be administered initially to the machine learning algorithm to get the correct direction towards the accurate value of weights and biases of the recurrent neural network. In a similar way, for the SINDy framework, it can give an initial estimate of the needed threshold levels to set up the sparsification of the matrix of coefficients.

Symbolic regression can be viewed as a form of machine learning. The data describes the time evolution of dynamical systems, where the aim is to infer and derive the ordinary differential equations (ODEs) that govern the dynamics. In SINDy, we select a family of functions to represent the terms of the desired ODEs, and derive the linear coefficient that explains how each function contributes to the dynamics [\citen{brunton2016sparse}]. 

In regression problems, the main contribution is provided merely by a few terms, making sparsity a critical element in the derivation. The sparsification is important as it removes negligible values from your resulting coefficient matrix. SINDy allows for the user to control the polynomial order, the number of variables, and whether to include sine and cosine functions in the library of functions $\Theta(X)$ before regression is applied. In the available SINDy framework, sparsification tolerances can be adjusted to control how sparse the result is. 

\begin{figure}[H]
\centering
  \includegraphics[scale=0.65]{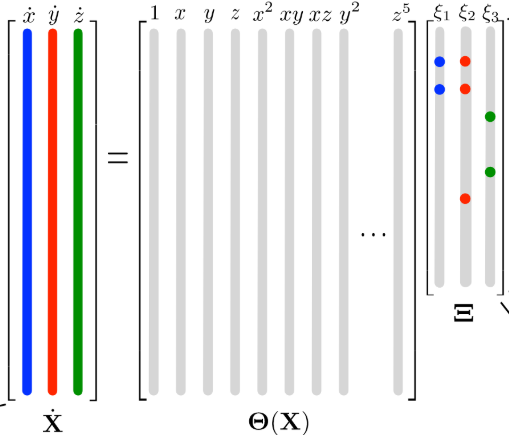}
  \caption{Diagram of SINDy constructing Dynamics}
  \label{fig:sindy_desc} 
\end{figure}

Conceptually, the regression is performing an estimation process. It tries estimating the matrix of coefficients $\mbf c$ that dictates the importance of each possible term (monomial) in the representation of the derivatives. The ODE is linear in the coefficients and can be written as [\citen{servadio2020nonlinear,servadio2025analytical}]
\begin{equation}
    \dot{\mbf X} = \boldsymbol \Theta(\mbf X) \mbf \Xi
\end{equation}
and visually represented in Fig. \ref{fig:sindy_desc}, where 
\begin{equation}
     \boldsymbol \Theta(\mbf X) = \begin{bmatrix}
        1 & \mbf x_1^{(1)} & \dots & \mbf x_n^{(1)} & (\mbf x_n^{(1)})^2 & \mbf x_1^{(1)}\mbf x_2^{(1)} & \dots \\
        1 & \mbf x_1^{(2)} & \dots & \mbf x_n^{(2)} & (\mbf x_n^{(2)})^2 & \mbf x_1^{(2)}\mbf x_2^{(2)} & \dots \\
        \vdots & \vdots & \dots & \vdots & \vdots & \vdots &  \dots & \\
        1 & \mbf x_1^{(N)} & \dots & \mbf x_n^{(N)} & (\mbf x_n^{(N)})^2 & \mbf x_1^{(N)}\mbf x_2^{(N)} & \dots \\
    \end{bmatrix}
\end{equation}
is the library of functions evaluated at the given dataset, one row per point. In the presented case, each point is the system state at the given time step, from $1$ to $N$. Therefore, the set of coefficients $\mbf \Xi$ for the polynomial approximation is estimated through least squares. Consider the following least squares cost function 
\begin{equation}
    \mathcal J (\mbf \Xi ) = \left(\dot{\mbf X} - \boldsymbol \Theta(\mbf X) \mbf \Xi \right)^T\left(\dot{\mbf X} - \boldsymbol \Theta(\mbf X) \mbf \Xi \right) \label{cost}
\end{equation}
the optimal estimate is obtained by setting the condition
\begin{equation}
    \dfrac{\partial \mathcal J (\mbf \Xi) }{\partial \mbf \Xi} = \mbf 0
\end{equation}
which leads to the solution
\begin{equation}
    \boldsymbol{\Xi} = (\boldsymbol{\Theta}(\mathbf X)^T\boldsymbol{\Theta}(\mathbf X))^{-1} \boldsymbol{\Theta}(\mathbf X)^T \dot{\mathbf{X}}
\end{equation}
However, SINDy encourages sparsity in $\mbf \Xi$, meaning that the cost function in Eq. \eqref{cost} has a penalizing term in the entries of $\mbf \Xi$ [\citen{brunton2016sparse}]. At the end, the estimated ODEs of the dynamics that best fit the data, in a least squares sense, are obtained via
\begin{equation}
    \dot{\mbf x} = \mbf f (\mbf x) = \mbf \Xi^T (\boldsymbol \Theta(\mbf x))^T
\end{equation}

An overview of the least square method is given in Fig. \ref{fig:sindy_desc}, where the $\dot{\mathbf{X}}$ represents the derivatives of the state, evaluated numerically via finite differences,  $\boldsymbol{\Theta}(\mathbf X)$ is the training matrix of the dataset at different time steps, and $\boldsymbol{\Xi}$ is the coefficient matrix, which fits the data into a set of polynomial ordinary differential equations. The sparsity of SINDy ensures that coefficients that are too small are set to null, moving the contributions to the other terms. 

The goal is to apply the SINDy algorithm to reproduce the dynamical relationships similar to the MOCAT-SSEM. The strengths of the SINDy approach over the current source-sink model are its ability to process noisy data and simplify dynamics through sparse regression, which is important with large volumes of turbulent data. Even in cases where data has little to no noise, the algorithm is able to derive dynamics of small to moderately sized systems in a reasonable amount of time.

As previously stated, the tuning of the model parameters is a major challenge. Thankfully, the MOCAT-MC tool allows for the active satellite population to be kept around a target value for the duration of the simulation. The sparsification tolerance cannot be set to too low a value as to introduce coefficients to the active satellite. We selected the polynomial library, with maximum order ranging from 1 to 3. Unfortunately, due to the complexity of the application and its large dimensionality, higher orders could not be selected. However, the second-order SINDy evaluation provides a fair and accurate comparison with the current model, as the MOCAT-SSEM equations of motion resample a second-order polynomial ODE.

\subsection{The Long-Short Term Memory (LSTM) Neural Network Approach }\label{sec:lstm}
The LSTM is a variation of the vanilla Recurring Neural Network (RNN), generally applied when given sequential data and a time series, such as the LEO space population evolution in time. The main advantage with respect to RNN consists in avoiding the problem of gradient vanishing, a common issue for long time series where multiple multiplications by weights smaller than unity make the gradient impossible to track numerically.

When developing an LSTM RNN model to predict LEO orbital population data provided by the MOCAT-MC model, the ultimate goal is to be able to replace the MOCAT-SSEM counterpart as a whole with the developed LSTM model. Some benefits of implementing the LSTM model over the MOCAT-SSEM model include: the potential for more accurate predictions, especially in regards to any changes in past data that the Source-Sink model would not be able to predict, more predictive flexibility in that the model could be re-trained over a different set of MOCAT data if necessary, and a wider input range, in that the LSTM model can train and predict on any set of MOCAT data input into it accurately. The development of this LSTM network was greatly influenced by the network developed by Klinkachorn et. al. [\citen{klinkachornevaluating}], which network has been used as an initial structure for the LSTM developed in this work. 

\begin{figure}[h]
\centering
  \includegraphics[scale=0.45]{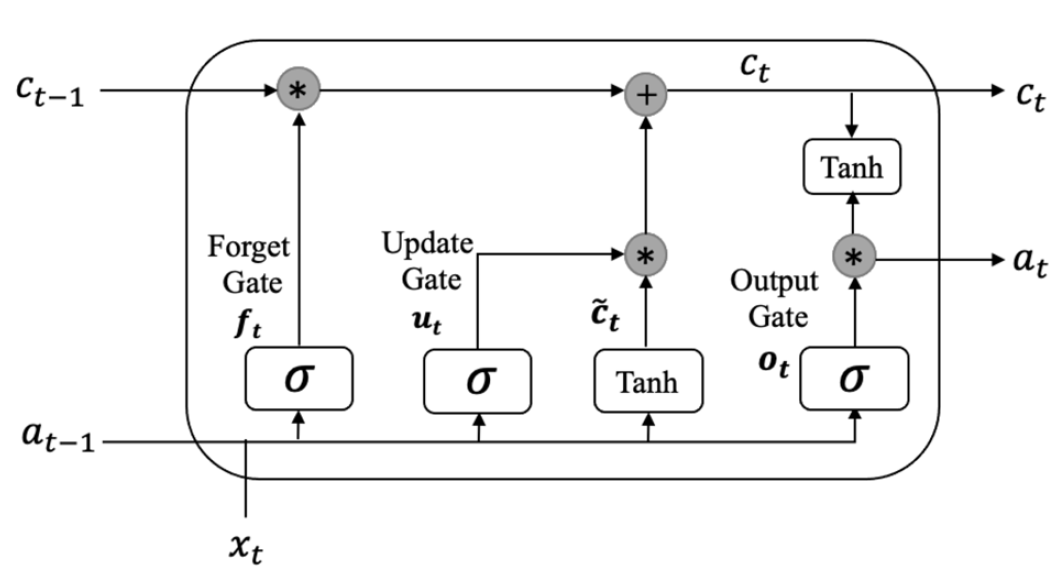}
  \caption{Internal Structure of an LSTM cell. Figure from [\citen{nguyen2022accurate}].}
  \label{fig:lstm} 
\end{figure}

Long Short‑Term Memory (LSTM) networks extend recurrent neural networks by introducing an explicit memory cell whose content can be controlled through multiplicative gates, allowing the model to learn long‑range dependencies without suffering severely from vanishing or exploding gradients. At each time step $t$, the LSTM cell receives an input vector $\mbf x_t$ to be merged with the previous hidden state as well as the previous cell (memory) state. Internally, three gate activations, i.e., forget, input, and output, visualized in Fig. \ref{fig:lstm}, modulate information flow. The idea is that, at each run of the LSTM architecture, new information (input) is merged with the memory from training from previous data, to provide the output and a short and long-term memory (hidden) state of the system. 

In an LSTM cell, Fig.~\ref{fig:lstm}, at time step \(t\) the network maintains a cell state \(c_{t}\) and a hidden (output) activation \(a_{t}\).  Given the new input \(x_{t}\) and the previous hidden state \(a_{t-1}\), three gates control the flow of information into and out of the cell. First, the forget gate, \(f_{t}\), decides which components of the previous cell state, \(c_{t-1}\), should be retained or discarded:
\begin{equation}
    f_{t} = \sigma\bigl(W_{f}\,[\,x_{t};\,a_{t-1}\,] + b_{f}\bigr)
\end{equation}
where \(\sigma(\cdot)\) is the logistic sigmoid function and \([\,x_{t};\,a_{t-1}\,]\) denotes vector concatenation.  The element‐wise product \(f_{t}\odot c_{t-1}\) propagates the retained portion of the old state. Next, the update gate, \(u_{t}\), (often called the ``input gate”) and the candidate state, \(\tilde{c}_{t}\), jointly determine how much new information to write into the cell:
\begin{align}
    \tilde{c}_{t} &= \tanh\bigl(W_{c}\,[\,x_{t};\,a_{t-1}\,] + b_{c}\bigr) \\
    u_{t} &= \sigma\bigl(W_{u}\,[\,x_{t};\,a_{t-1}\,] + b_{u}\bigr)
\end{align}
The product \(u_{t}\odot\tilde{c}_{t}\) is then added to the retained old state to yield the new cell state
\begin{equation}
   c_{t} = f_{t}\odot c_{t-1} \;+\; u_{t}\odot \tilde{c}_{t}.
\end{equation}
Finally, the output gate, \(o_{t}\), regulates which parts of the updated cell state become the hidden activation:
\begin{align}
    o_{t} &= \sigma\bigl(W_{o}\,[\,x_{t};\,a_{t-1}\,] + b_{o}\bigr) \\
    a_{t} &= o_{t}\odot \tanh(c_{t})
\end{align}
Altogether, the matrices \(W_{f},W_{u},W_{c},W_{o}\) and bias vectors \(b_{f},b_{u},b_{c},b_{o}\) are learned to optimize the gating behavior, enabling the LSTM cell to preserve, update, and emit information selectively over long sequences.

During the development of the LSTM network, the model was trained on two different sets of data, referred to as Case 1 and Case 2, respectively. These two different datasets are the total number of S, D, and N objects in all 36 atmospheric shells, and the population of each object type within each individual shell. When developing the LSTM network, one of the greatest challenges is tuning the parameters to obtain the best predictive scenario of the data. When tuning these parameters, the main issue is optimizing the predictive scenario to be both as accurate and as computationally efficient as possible. When performing this optimization, the parameters that were tuned were the number of hidden LSTM layers within the model, the number of hidden nodes within each layer, the training data ratio, activation function selection, loss function selection, sequence length, batch size, and number of training epochs. These parameters were tuned and changed throughout the development of the LSTM network in order to best optimize the network: the final selection for each case is reported in Table \ref{tab: lstm_config}.
\begin{table}[htbp]
\centering
\begin{tabular}{|l|c|c|}
\hline
\textbf{Parameter} & \textbf{CASE 1} & \textbf{CASE 2}\\
\hline
Number of Input Features          & 3   & 108\\ 
Number of LSTM Layers            & 2   & 1\\ 
Number of Hidden Units per Layer & 64/32   & 20\\ 
Sequence Length (time steps)     & 50  & 50\\ 
Output Size                      & 3 & 108\\ 
Activation Function (internal)     & tanh & tanh\\ 
Activation Function (recurrent)     & sigmoid & sigmoid\\ 
Activation Function (output)     & linear & linear\\ 
Loss Function                    & MSE & MSE\\ 
Optimizer                        & Adam & Adam\\ 
Batch Size                       & 64 & 1\\ 
Number of Epochs                 & 100 & 50\\ 
\hline
\end{tabular}
\caption{LSTM Network Configuration}
\label{tab: lstm_config}
\end{table}

For Case 1 referenced in Table \ref{tab: lstm_config}, the model was trained on total S, D, and N population data across all 36 atmospheric shells from the first 1,000 MOCAT-MC simulations. This training was performed by separating the data from these 1,000 simulations into groups of 10 and taking the average of the data at each timestep within these groups, so that 100 averaged datasets were created. The model was then trained continuously on all 100 datasets to create a model that could predict the total S, D, and N population for any variation of the MOCAT-MC simulation. Parameters for this model were chosen to reflect the complexity of training on 100 different datasets, while also minimizing computational time. As shown in Table \ref{tab: lstm_config}, the model had an input and output size of 3, for training and predicting on S, D, and N populations. The model contained two LSTM layers, with hidden node counts of 64 and 32, respectively, in addition to a Dense output layer and a sequence length of 50, in order to capture the complexity of the dataset. \textit{Tanh}, \textit{sigmoid}, and \textit{linear} activation functions were selected for the internal, recurrent, and output activation functions, respectively. To ensure full training on each dataset, 100 epochs were used during training. However, in order to conserve computation time, early stopping with a patience setting of 5 epochs was employed, and a batch size of 64 was used.

For Case 2 referenced in Table \ref{tab: lstm_config}, the model was trained on S, D, and N population data from each of the individual 36 shells. For this case, the model was trained on a dataset that was taken from the average of each of the first 1,000 MOCAT-MC simulations. From Case 1 to Case 2, as seen in Table \ref{tab: lstm_config}, the input size of the model changed from 3 to 108 to account for each of the 36 shells for all S, D, and N objects. The number of LSTM layers was reduced to 1, and the number of hidden nodes was reduced to 20 to increase computation time, as the model was not trained on many different datasets. The batch size of the model training was reduced to 1 for accurate training on the dataset the model was trained on. Finally, the number of epochs was reduced from 100 to 50. However, early stopping was not used in Case 2, so the model could train for all 50 epochs. All results of the predictions of the LSTM model for both cases based on the data provided by the MOCAT-MC simulations will be discussed in the Results section.

%% file: 04_results.tex
\section{Numerical Results}
\label{sec:results}
The two separate approached have been compared to the dataset time history given to analyze how closely they match the high-fidelity dynamics.

\section{SINDy Model Results}
In the initial evaluation, SINDy has been applied to estimate the overall LEO population and to obtain a set of ODEs that propagate the total number of active satellites, derelicts, and debris that orbit Earth, without shell division. This would reduce the data to a 3-dimensional system. The goal is to reproduce the trend of the data and keep the error difference between the SINDy and MOCAT-MC results as low as possible. 

\begin{figure}[h]
    \centering
    \begin{subfigure}[b]{0.49\textwidth}
        \includegraphics[width=\textwidth]{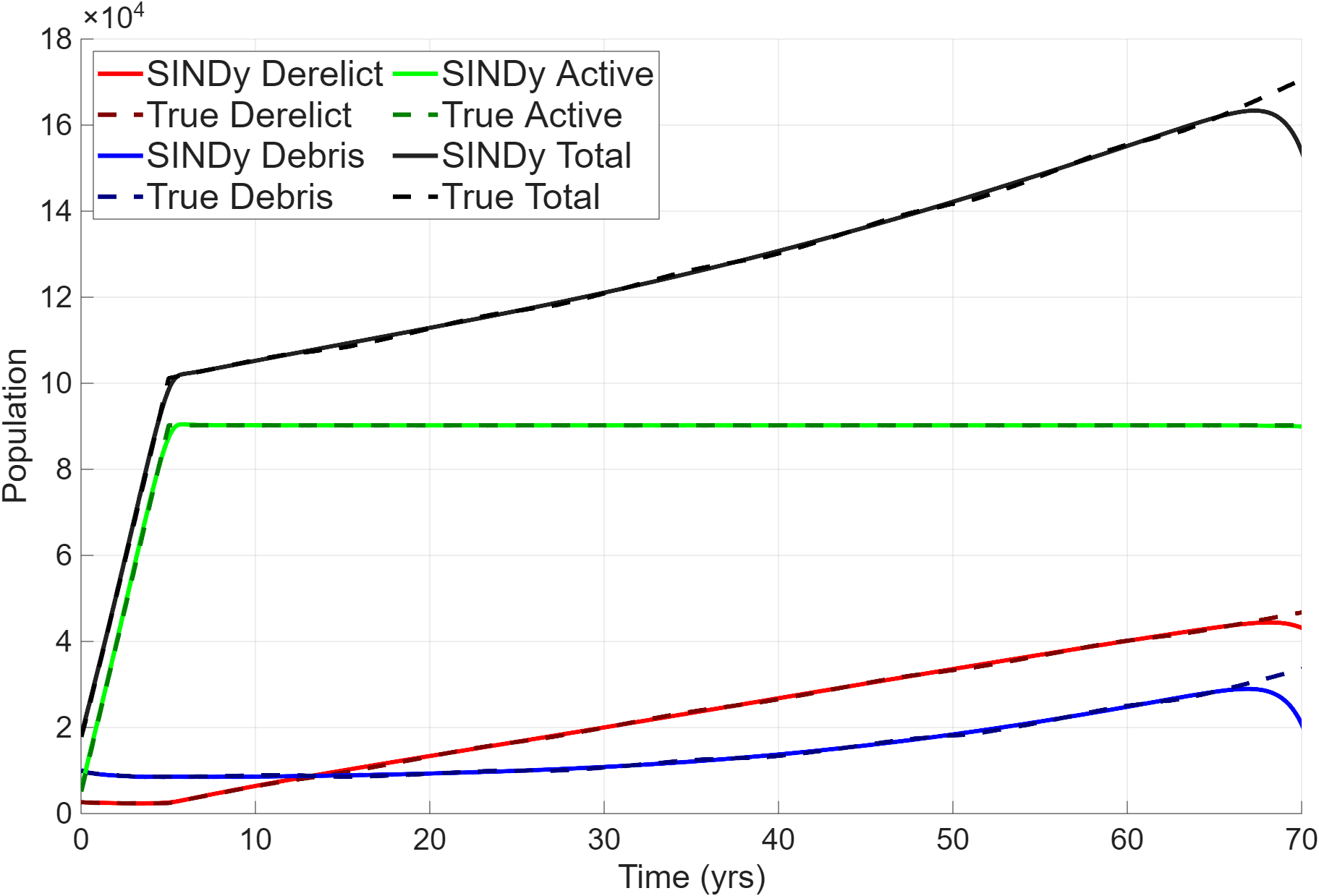}
        \caption{\footnotesize Population Data for Average}
        \label{fig: SINDy Average Data}
    \end{subfigure}
    \hfill
    \begin{subfigure}[b]{0.49\textwidth}
        \includegraphics[width=\textwidth]{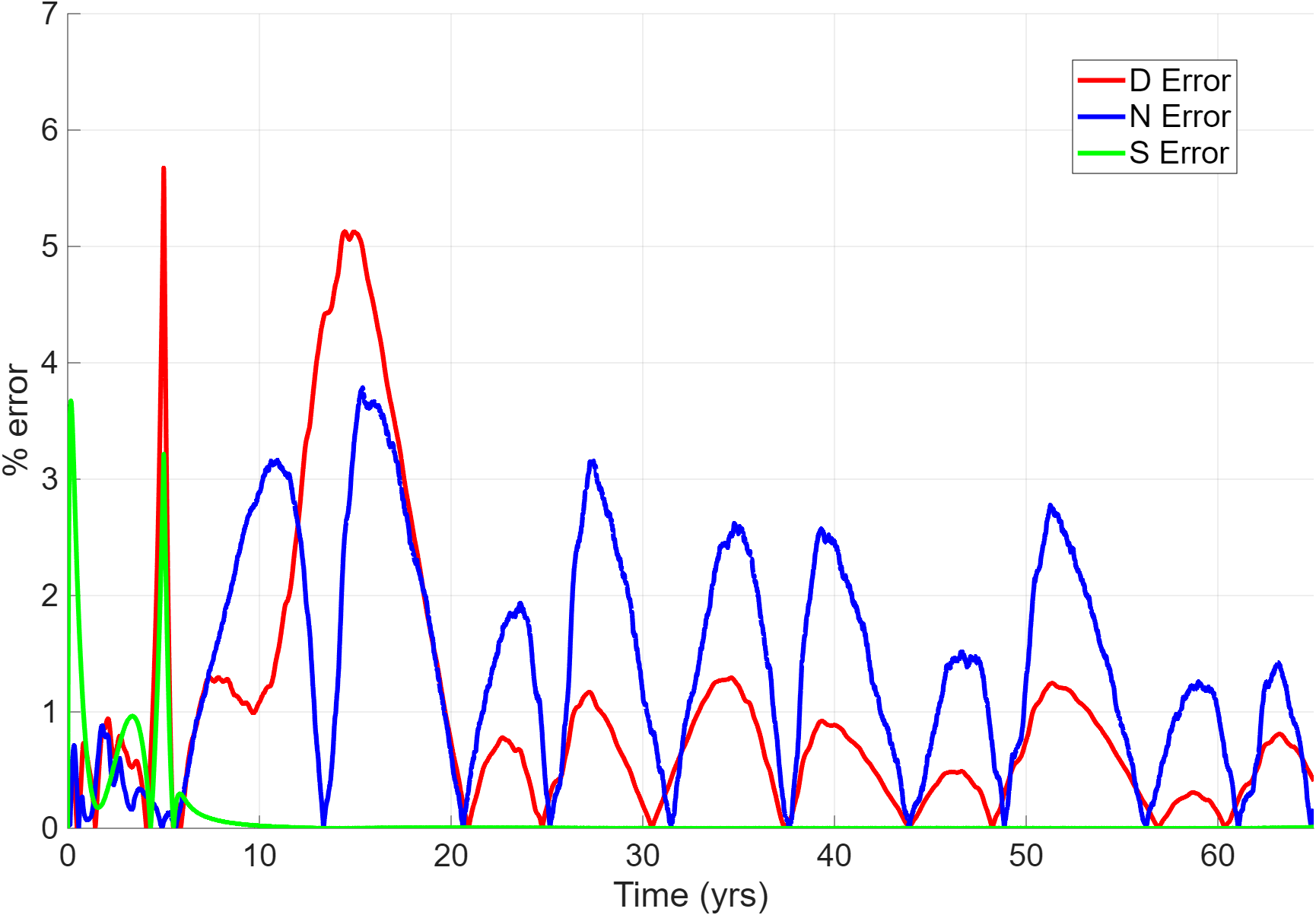}
        \caption{\footnotesize Percent Error for Average}
        \label{fig: SINDy Average Error}
    \end{subfigure}
    \caption{Dynamics To Source-Sink Simulation Comparison and error for Total LEO population From SINDy Training along with coefficients for dynamics}
    \label{fig: SINDy Average}
\end{figure}

The dynamics have been derived using a third-order, $n=3$, polynomial library for the library functions. Thus, the plots in Figure \ref{fig: SINDy Average} were generated using the sum of all satellites in LEO, divided by family. Figure \ref{fig: SINDy Average Data} compares the MC data with the numerically integrated polynomial dynamics derived via the SINDy approach, while Fig. \ref{fig: SINDy Average Error} shows the relative error, evaluated as a percentage between prediction and validation data.

It can be noted that the derived dynamics correctly model the gradient discontinuity in the number of active satellites. However, the derived ODEs break down around the year 70, as the population diverges. This is due to the accuracy of the entries in the coefficient matrix, where more decimals become of importance the longest it is desired to propagate the population. For a short-term evaluation, the model is extremely accurate, with error level setting below 3\%. Indeed, Fig. \ref{fig: SINDy Average Error} shows that the debris family has the highest error level, followed by the derelict curve. Both curves have a periodic pattern, behavior that is likely due to the period change in atmospheric density coming from the Sun's activity. With the polynomial library of functions, SINDy averages the oscillations with a smooth curve. In future developments, we will introduce coefficients with trigonometric basis functions at different frequencies to better estimated the true dynamics, increasing the computational complexity and burden but improving the model adaptation. 

\begin{figure}[H]
\centering
  \includegraphics[width=1\linewidth]{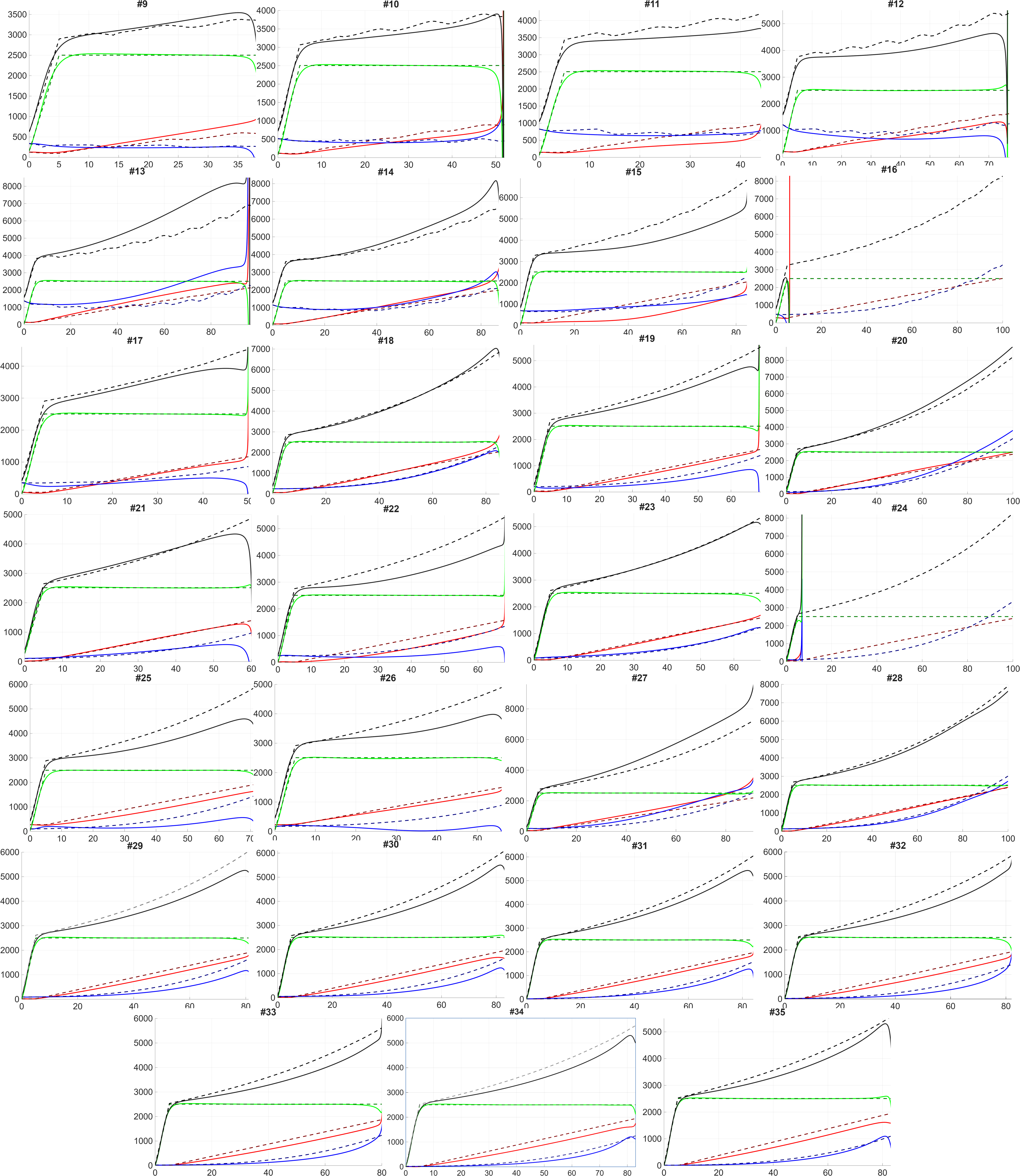}
  \caption{SINDy Single shell dynamics evaluation}
  \label{fig:all} 
\end{figure}

The following steps consist of obtaining a set of three-dimensional ODEs for each given shell of the MOCAT model. In MOCAT-SSEM, the dynamics of a shell are coupled just with the shell above and below, as derelicts drop from the drag effect, active satellites rise to their operational altitude, and satellites that have completed their mission re-enter following their PDM. That is, each orbital shell obtains information only from the shell above and below itself. Therefore, the SINDy approach and training have been applied in batches of three shells at a time, where the learning of the $i$th shells is performed with data from the shell $i-1$, $i$, and $i+1$, for a total of  9 dimensions. Order 2 has been selected to match the MOCAT-SSEM polynomial model, but higher orders with larger libraries are planned to be applied in the future after accessing High-Performance Computers (HPCs).   

When training on individual shells, an additional consideration is due. Low orbit shells, such as the ones covering altitudes between 200 km and 500 km, have extremely sparse data with many null entries. These shells are barely populated, as most of their RSOs are just traveling through and reentering due to the high drag effects. Therefore, for these shells, the SINDy dynamics prediction is poor in performance, as the data itself is flawed. Figure \ref{fig:all} reports the SINDy dynamics estimation comparison for all orbital shells from 9 to 35 (as shell 36 would need data from a non-existing shell to be trained). The SINDy dynamics is accurate and correct for the full family, with validity spanning from year 40 to year 100, the last one. Exceptions are shells 16 and 24, where the population breaks right at the gradient discontinuity. The training has been performed with the same coefficient threshold for all the shells, and these two require particular attention and tuning to obtain an accurate result. 

The figure shows the promising results and learning that can be obtained via the SINDy methodology and that the LEO population forecasting via the new derived polynomial dynamics is accurate for the short-term propagation, which is usually of interest, and for every shell but two exceptions.  

\section{LSTM Model Results}
Using the LSTM training and predicting methods as described earlier, prediction data were obtained for the last 20 years of the 100-year time period over which MOCAT-MC data were simulated. As previously mentioned, both the total S, D, and N object population data combined across all atmospheric shells and the object population data for each individual shell were trained on and predicted. 

\subsection{Case 1}
\begin{figure}[h]
    \centering
    \begin{subfigure}[b]{0.49\textwidth}
        \includegraphics[width=\textwidth]{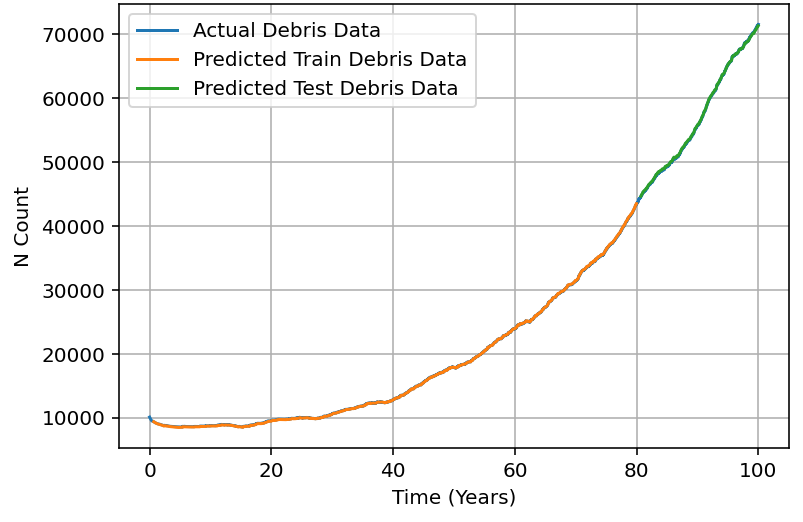}
        \caption{\footnotesize N Object Population Data}
        \label{fig: LSTM Augmented N Data}
    \end{subfigure}
    \hfill
    \begin{subfigure}[b]{0.49\textwidth}
        \includegraphics[width=\textwidth]{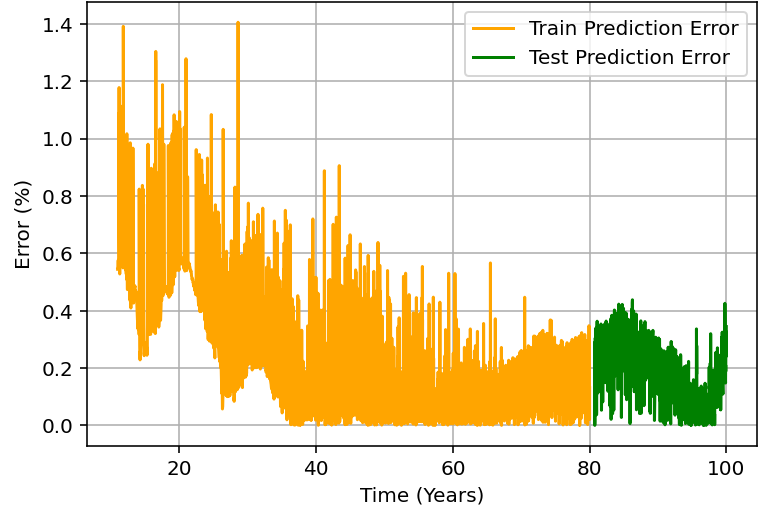}
        \caption{\footnotesize N Object Percent Error}
        \label{fig: LSTM Augmented N Error}
    \end{subfigure}
    \caption{Combined Atmospheric Shell N Object Data and Error for LSTM Predictions}
    \label{fig: LSTM Augmented N}
\end{figure}

\begin{figure}[h]
    \centering
    \begin{subfigure}[b]{0.49\textwidth}
        \includegraphics[width=\textwidth]{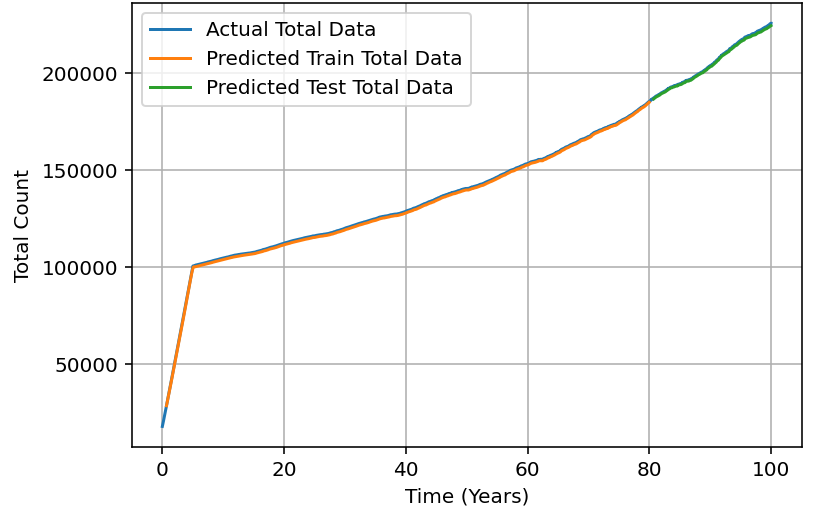}
        \caption{\footnotesize Total Object Population Data}
        \label{fig: LSTM Augmented Total Data}
    \end{subfigure}
    \hfill
    \begin{subfigure}[b]{0.49\textwidth}
        \includegraphics[width=\textwidth]{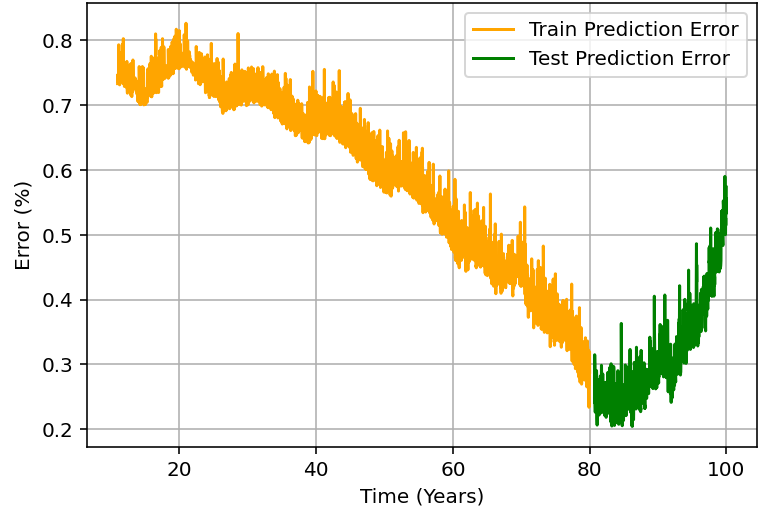}
        \caption{\footnotesize Total Object Error Percentage}
        \label{fig: LSTM Augmented Total Error}
    \end{subfigure}
    \caption{Combined Atmospheric Shell Total Object Data and Error for LSTM Predictions}
    \label{fig: LSTM Augmented Total}
\end{figure}

For the total population data, the prediction was created from a model trained on 100 different simulations, varying slightly due to the Monte Carlo nature of the MOCAT-MC. Shown below in Figures \ref{fig: LSTM Augmented S}, \ref{fig: LSTM Augmented D}, \ref{fig: LSTM Augmented N}, and \ref{fig: LSTM Augmented Total} are plots of the predicted total object population data compared to the corresponding training and validation values. The data is broken into each object type separately, with an additional plot for the total number of objects overall, describing the overall LEO space environment orbit capacity. Moreover, for each prediction figure, there is a corresponding error analysis has been performed and reported. This error is defined as the percent value for the ratio of the difference between the LSTM predicted data and the real data from the high-fidelity model. These error plots omit the first 750 time steps to avoid plotting large error percentages at the beginning of training.

Figure \ref{fig: LSTM Augmented S Data} shows the active satellite, S, number over time, both from training data and predicted from the LSTM. The MOCAT-MC simulation is set to keep the number of satellites orbiting constant, with a constant launch rate to replenish satellites that complete their mission. Therefore, the S line is constant after reaching its steady state value. The LSTM has no problem correctly predicting this behavior, as shown in Fig. \ref{fig: LSTM Augmented S Error} for the relative error. 

\begin{figure}[h]
    \centering
    \begin{subfigure}[b]{0.49\textwidth}
        \includegraphics[width=\textwidth]{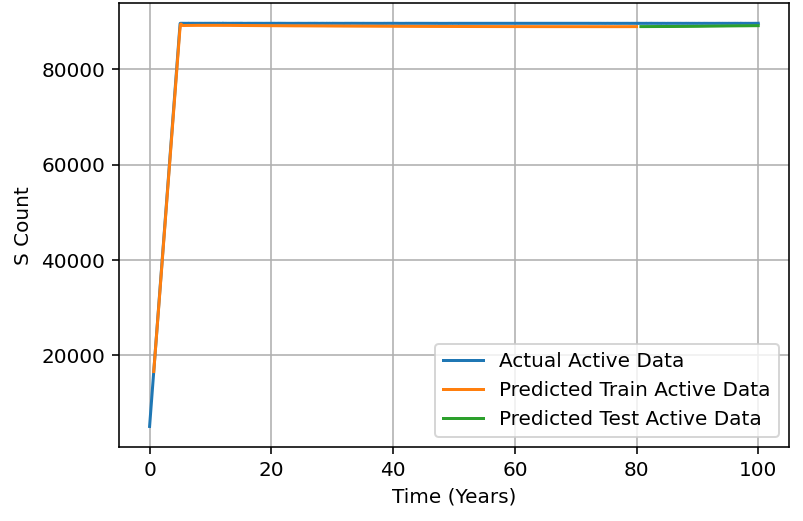}
        \caption{\footnotesize S Object Population Data}
        \label{fig: LSTM Augmented S Data}
    \end{subfigure}
    \hfill
    \begin{subfigure}[b]{0.49\textwidth}
        \includegraphics[width=\textwidth]{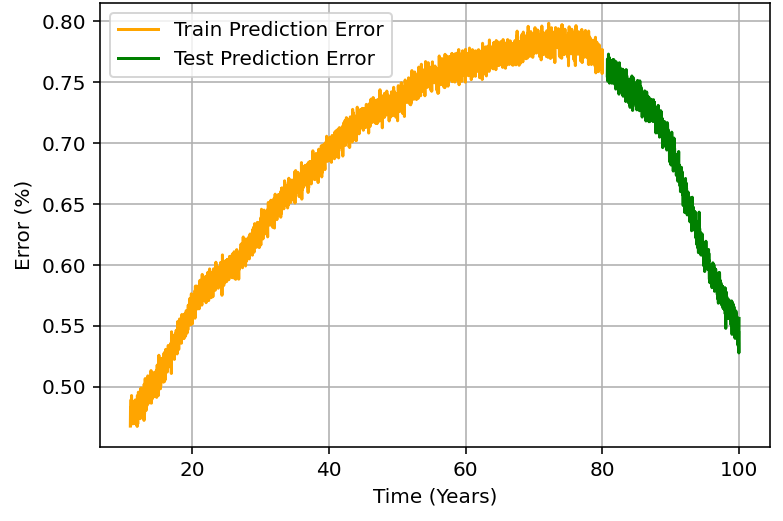}
        \caption{\footnotesize S Object Percent Error}
        \label{fig: LSTM Augmented S Error}
    \end{subfigure}
    \caption{Combined Atmospheric Shell S Object Data and Error for LSTM Predictions}
    \label{fig: LSTM Augmented S}
\end{figure}

Figure \ref{fig: LSTM Augmented D Data} shows the derelict, D, number over time, which increases linearly each year. In a similar way, Fig. \ref{fig: LSTM Augmented D Error} shows the relative error of the LSTM on both the training and validation datasets. As expected, the error quickly reduces from initial high levels (not reported in the figure for visualization purposes), reducing drastically with time. After 80 years, in the 20-year prediction, the error keeps on acceptable levels for the first 10 years, and then starts increasing, as expected from the LSTM mathematics, where the farthest in the prediction of the state is assumed, the less accurate it becomes. However, the overall error value keeps below the 1\% line. 

\begin{figure}[h]
    \centering
    \begin{subfigure}[b]{0.49\textwidth}
        \includegraphics[width=\textwidth]{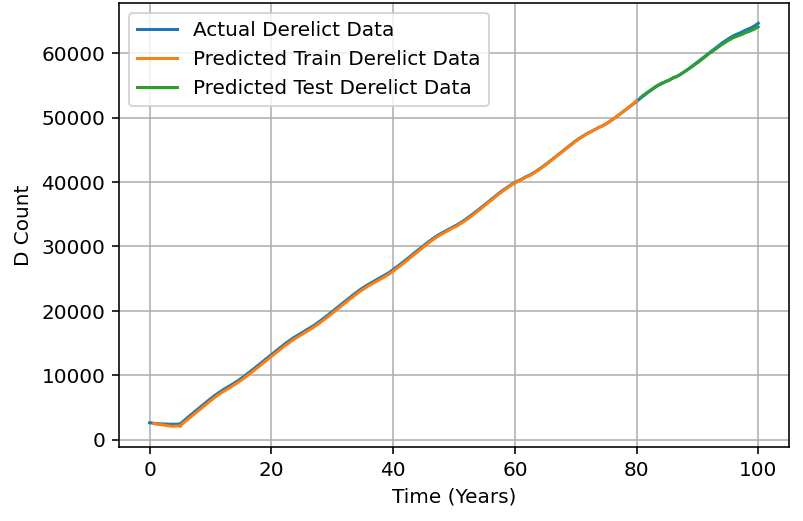}
        \caption{\footnotesize D Object Population Data}
        \label{fig: LSTM Augmented D Data}
    \end{subfigure}
    \hfill
    \begin{subfigure}[b]{0.49\textwidth}
        \includegraphics[width=\textwidth]{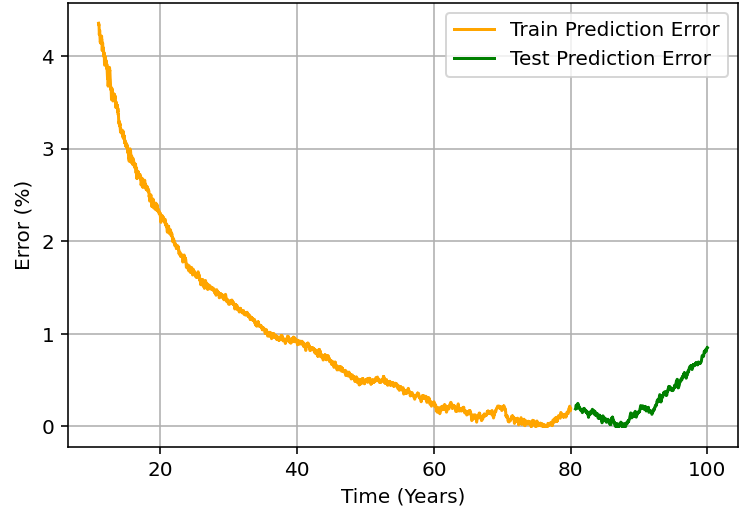}
        \caption{\footnotesize D Object Percent Error}
        \label{fig: LSTM Augmented D Error}
    \end{subfigure}
    \caption{Combined Atmospheric Shell D Object Data and Error for LSTM Predictions}
    \label{fig: LSTM Augmented D}
\end{figure}

Figure \ref{fig: LSTM Augmented N Data} shows the debris, N, number over time, which increases exponentially each year. The debris dynamics is the hardest to estimate, but the LSTM is able to accurately predict its behavior, as shown in Fig.  \ref{fig: LSTM Augmented N Error}. With respect to the previous error figure, the N data is the most noise, due to the stochastic collision and fragmentation events that create discontinuities and jumps in the data. Therefore, the relative error of N is a band rather than a line, due to the noisy signal given as training. However, the overall trend still shows very accurate prediction with error rates below 0.5\%.

Lastly, the overall number of RSOs in LEO is reported in Fig. \ref{fig: LSTM Augmented Total Data}, with relative error plotted in Fig. \ref{fig: LSTM Augmented Total Error}. This graph validates the use of the LSTM to give a quick initial prediction of the future LEO orbital capacity whenever requesting launch rate changes or studying quick action/consequences effects on the orbital shells without having to perform the computational expensive high-fidelity MOCAT-MC propagation. The error plot shows that the longer the training time series, the more accurate the LSTM becomes at providing the initial prediction. However, the prediction (in green) becomes less accurate the further in time it gets from the last training data point. 

From the data above in Figures \ref{fig: LSTM Augmented S}, \ref{fig: LSTM Augmented D}, \ref{fig: LSTM Augmented N}, and \ref{fig: LSTM Augmented Total}, it can be seen that the LSTM model predicted the MOCAT-MC object total population data very well, with the maximum prediction percent error being less than 1\% for the D object prediction in Figure \ref{fig: LSTM Augmented D Error}. This data is very encouraging, and is what inspired training the LSTM model on the individual atmospheric shell MOCAT-MC data for each object type as well.

\subsection{Case 2}
\begin{figure}[h]
    \centering
    \begin{subfigure}[b]{0.49\textwidth}
        \includegraphics[width=\textwidth]{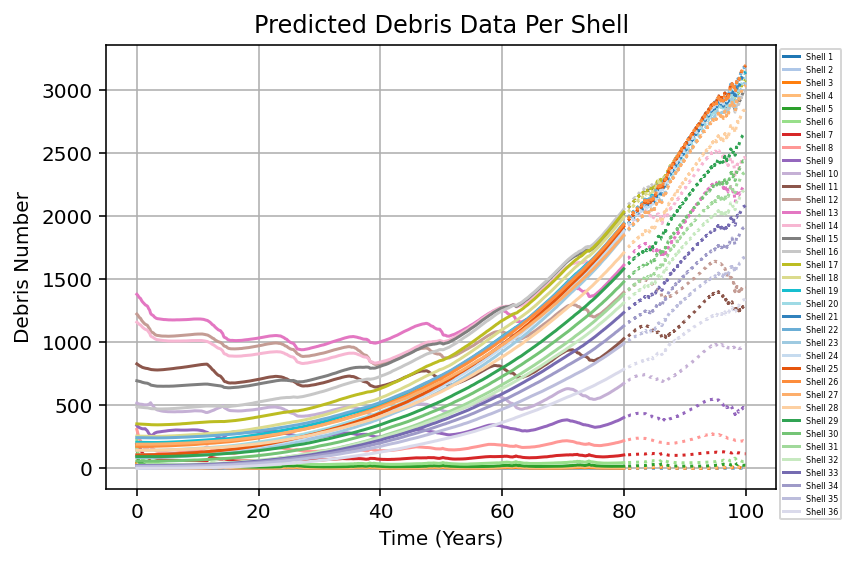}
        \caption{\footnotesize N Object Population Data}
        \label{fig: LSTM Individual N Data}
    \end{subfigure}
    \hfill
    \begin{subfigure}[b]{0.49\textwidth}
        \includegraphics[width=\textwidth]{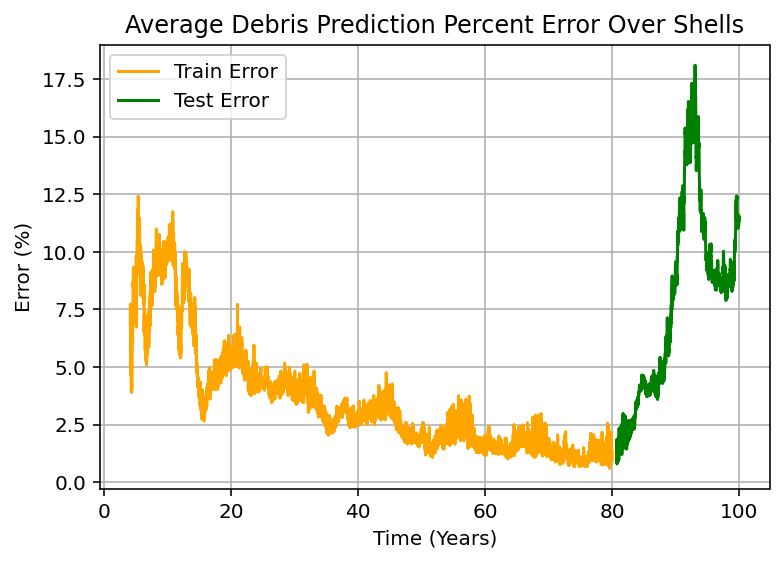}
        \caption{\footnotesize N Object Percent Error}
        \label{fig: LSTM Individual N Error}
    \end{subfigure}
    \caption{Individual Atmospheric Shell N Object Data and Error for LSTM Predictions}
    \label{fig: LSTM Individual N}
\end{figure}

\begin{figure}[h]
    \centering
    \begin{subfigure}[b]{0.49\textwidth}
        \includegraphics[width=\textwidth]{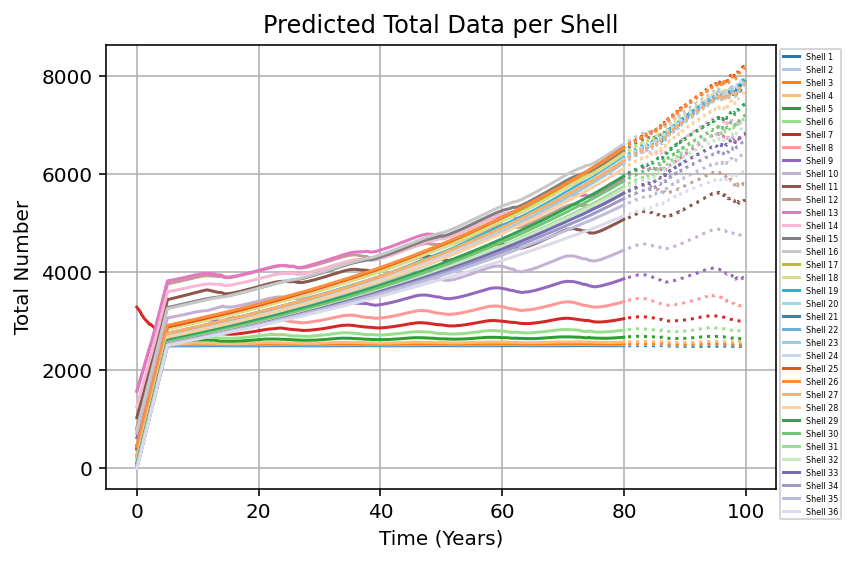}
        \caption{\footnotesize Total Object Population Data}
        \label{fig: LSTM Individual Total Data}
    \end{subfigure}
    \hfill
    \begin{subfigure}[b]{0.49\textwidth}
        \includegraphics[width=\textwidth]{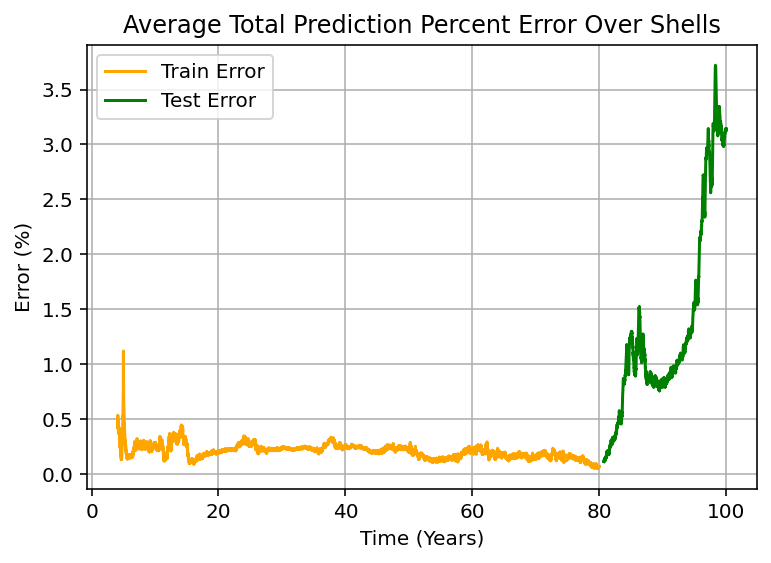}
        \caption{\footnotesize Total Object Percent Error}
        \label{fig: LSTM Individual Total Error}
    \end{subfigure}
    \caption{Individual Atmospheric Shell Total Object Data and Error for LSTM Predictions}
    \label{fig: LSTM Individual Total}
\end{figure}

As mentioned previously, the shell individual data was predicted from an LSTM model trained on the time step average of the first 1,000 MOCAT-MC simulations. Shown below in Figures \ref{fig: LSTM Individual S}, \ref{fig: LSTM Individual D}, \ref{fig: LSTM Individual N}, and \ref{fig: LSTM Individual Total} are plots of the predicted individual atmospheric shell MOCAT-MC simulation data compared to the corresponding training data and real values. The data is once again broken into each object type separately, with an additional plot for the total number of objects in each atmospheric shell overall. The error plots corresponding to each object type population prediction show the average of the error plot values for each of the individual shells at each time step. The error values in Figures \ref{fig: LSTM Individual S Error}, \ref{fig: LSTM Individual D Error}, \ref{fig: LSTM Individual N Error}, and \ref{fig: LSTM Individual Total Error} were calculated similarly to those in Figures \ref{fig: LSTM Augmented S Error}, \ref{fig: LSTM Augmented D Error}, \ref{fig: LSTM Augmented N Error}, and \ref{fig: LSTM Augmented Total Error}. It is important to mention that the MOCAT-MC simulation that generated the dataset was set up to keep a constant of 2500 active satellites in each shell, with a replenishing launch rate. Thus, the S lines among shells overlap with those values due to the simulation parameter setting. 

\begin{figure}[h]
    \centering
    \begin{subfigure}[b]{0.49\textwidth}
        \includegraphics[width=\textwidth]{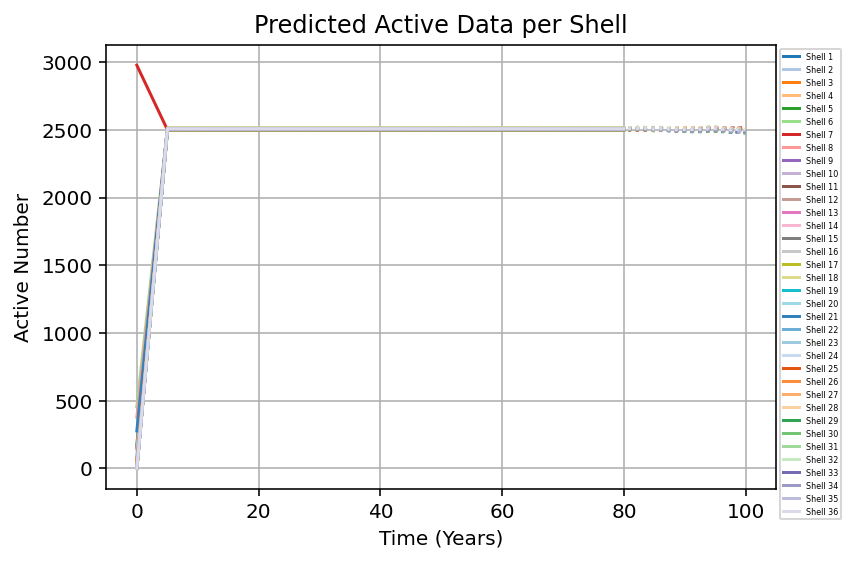}
        \caption{\footnotesize S Object Population Data}
        \label{fig: LSTM Individual S Data}
    \end{subfigure}
    \hfill
    \begin{subfigure}[b]{0.49\textwidth}
        \includegraphics[width=\textwidth]{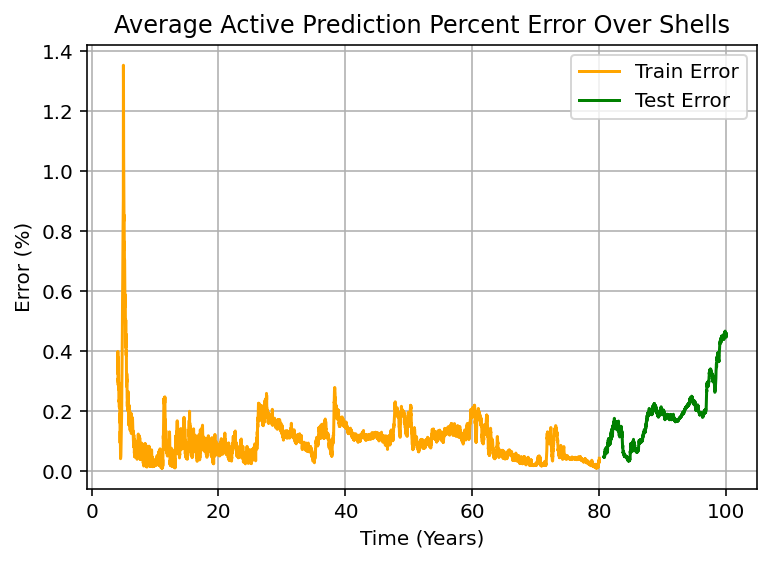}
        \caption{\footnotesize S Object Percent Error}
        \label{fig: LSTM Individual S Error}
    \end{subfigure}
    \caption{Individual Atmospheric Shell S Object Data and Error for LSTM Predictions}
    \label{fig: LSTM Individual S}
\end{figure}

\begin{figure}[h]
    \centering
    \begin{subfigure}[b]{0.49\textwidth}
        \includegraphics[width=\textwidth]{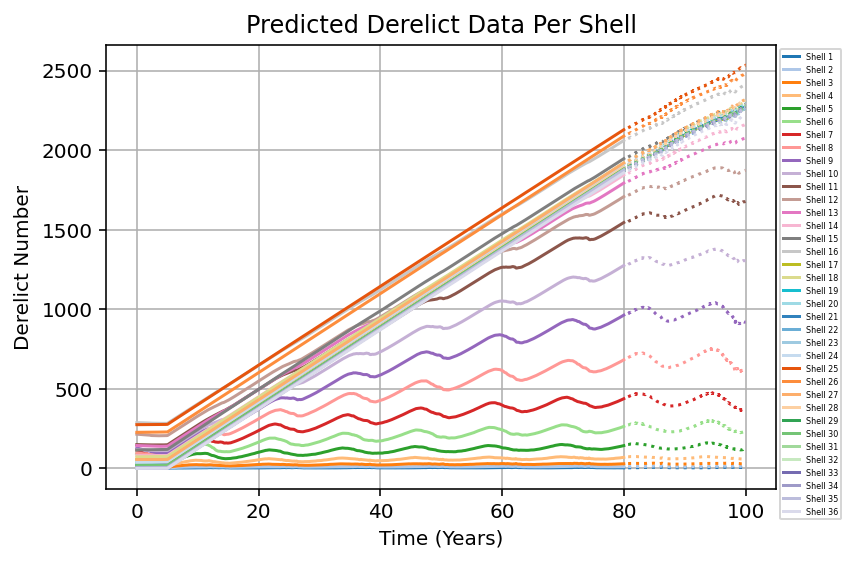}
        \caption{\footnotesize D Object Population Data}
        \label{fig: LSTM Individual D Data}
    \end{subfigure}
    \hfill
    \begin{subfigure}[b]{0.49\textwidth}
        \includegraphics[width=\textwidth]{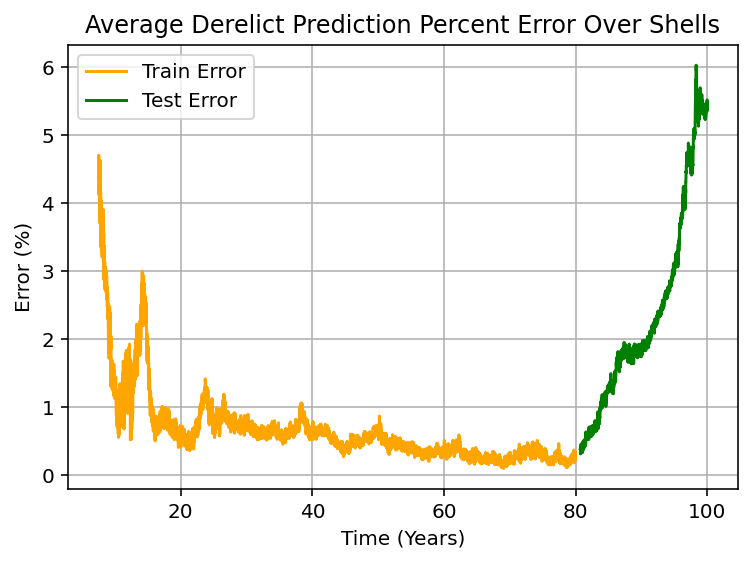}
        \caption{\footnotesize D Object Percent Error}
        \label{fig: LSTM Individual D Error}
    \end{subfigure}
    \caption{Individual Atmospheric Shell D Object Data and Error for LSTM Predictions}
    \label{fig: LSTM Individual D}
\end{figure}

From the data above in Figures \ref{fig: LSTM Individual S}, \ref{fig: LSTM Individual D}, \ref{fig: LSTM Individual N}, and \ref{fig: LSTM Individual Total}, it can be seen that, although the error for the individual atmospheric shell predictions is higher than the error for the total object population predictions for most object types, the LSTM model still predicted the MOCAT-MC simulation data fairly well for each individual shell. From Figures \ref{fig: LSTM Individual S Error} and \ref{fig: LSTM Individual Total Error}, it can be seen that the maximum average percent error for the S object population and the total population of objects, respectively, is about 0.5\% and 3.75\%. Both of these error values are low, and it can be concluded that the LSTM model trained very well on these datasets. From Figure \ref{fig: LSTM Individual D Error}, it can be seen that the maximum average percent error for the D object population is about 6\%. This error value is fairly low, and it can be concluded that the LSTM model trained fairly well on the D object population dataset. Finally, from Figure \ref{fig: LSTM Individual N Error}, it can be seen that the maximum average percent error for the N object population increases drastically, as expected, following the sinusoidal pattern of the atmospheric density. Although improvements to the model in regards to the prediction of the N object population for each individual atmospheric shell have already begun, this error value is still relatively low, and it can be concluded that the LSTM model trained acceptably on the N object population dataset for each individual atmospheric shell.

In the future, improvements will continue to be made to the model to improve its predictive capability for the individual atmospheric shell data from the MOCAT-MC simulations, specifically for the N object population dataset. One specific improvement that will be made will be that methods will be researched regarding increasing the sequence length for the LSTM hidden layer of the network to attempt to improve the long-term trend predictive capability of the model. In addition, the model will be trained on multiple different MOCAT-MC datasets for the individual shell data, like the total population data was. This will both help with the improvement of the LSTM model and provide insight into how the model can be used to replace the MOCAT-SSEM in the future.

%% file: 05_significance.tex
\section{Conclusions}
\label{sec:significance}
In a world where LEO traffic is increasingly in demand and necessary for many everyday functions, being able to accurately monitor and predict the population of objects within LEO is essential. By being able to predict this population, future missions to LEO and beyond can be planned to ensure the safety of the LEO space environment. The MOCAT model will play a very critical role in this effort by accurately predicting the future population of active satellites, derelict satellites, and debris objects in different shells of LEO. Every effort that is made to improve the accuracy of the MOCAT in turn benefits the future of space flight by creating an even more reliable and accurate future prediction of the object population of LEO. Both the SINDy and LSTM approaches discussed in this paper offer a data-driven push to increase model accuracy by replacing the differential equation from the source-sink MOCAT model. 

The two approaches presented a valid substitute to the source-sink ODE approach. The SINDy evaluation provided a new set of polynomial ODE trained directly from he high-fidelity dataset. It showed an accurate overlapping for the short-term prediction, while having difficulties for the long-term simulations. In future developments, a more powerful training embedded in high-performance computers will provide more accurate results, where the library of functions can be expanded to higher orders and be enhanced with the introduction of trigonometric functions at different frequencies. 

On the other hand, the LSTM prediction shows a correct evolution of the population forecasting. While powerful, the LSTM is more computationally heavy and requires to be given an initial, even if short, set of data in the time series to accurately provide the future predicted states. In future developments, the network will be improved by enhancing its structure (in terms of nodes and hidden layers) and by additional training after data augmentation techniques.